\pdfoutput=1
\documentclass[11pt]{article}

\usepackage[final]{acl}

\usepackage{times}
\usepackage{algpseudocode}
\usepackage{xcolor}
\usepackage[utf8]{inputenc}
\usepackage{latexsym}
\usepackage{amssymb}
\usepackage{mathtools} 
\usepackage{underscore}
\usepackage{array}
\usepackage{graphicx}
\usepackage{amsmath}
\usepackage{float}
\usepackage{multirow}
\usepackage{algorithm2e}
\usepackage[T1]{fontenc}
\usepackage[utf8]{inputenc}

\usepackage{microtype}

\usepackage{inconsolata}

\begin{document}
\title{SCaLAR NITK at SemEval-2024 Task 5: Towards Unsupervised Question Answering system with Multi-level Summarization for Legal Text}
\author{
  \begin{tabular}[t]{c@{\hspace{1cm}}c@{\hspace{1cm}}c}
    M Manvith Prabhu$^*$ & Haricharana Srinivasa$^\dagger$ &  Anand Kumar M$\S$ \\
  \end{tabular} \\
  Department of Electronics and Communication $^*$, \\ 
  Department of Chemical Engineering $^\dagger$, \\ 
  Department of Information Technology $\S$, \\
  National Institute of Technology Karnataka (NITK), Surathkal - 575025, India\\
 \texttt{\{manvithprabhu.211ec228, sharicharana.211ch024, m\_anandkumar\}@nitk.edu.in}
}

\maketitle
\begin{abstract}
This paper summarizes Team SCaLAR's work on SemEval-2024 Task 5: Legal Argument Reasoning in Civil Procedure. To address this Binary Classification task, which was daunting due to the complexity of the Legal Texts involved, we propose a simple yet novel similarity and distance-based unsupervised approach to generate labels. Further, we explore the Multi-level fusion of Legal-Bert embeddings using ensemble features, including CNN, GRU and LSTM. To address the lengthy nature of Legal explanation in the dataset, we introduce T5-based segment-wise summarization, which successfully retained crucial information, enhancing the model's performance. Our unsupervised system witnessed a 20-point increase in macro F1-score on the development set and a 10-point increase on the test set, which is promising given its uncomplicated architecture.
\end{abstract}

\section{Introduction}

The Domain of Law demands sheer expertise and experience for a human to master, but it takes much more to teach a machine the same. Legal NLP (Zhong et al., 2020) is advancing at a rapid pace, and the advent of Transformers \cite{46201} has widened the prospects of research in this area. However, the intricate nature of Legal Texts and the underlying complex relationships between entities make it difficult even for state-of-the-art Language models like BERT \cite{devlin-etal-2019-bert} to capture the details effectively. To advance our understanding of the reasoning ability of LLMs in the legal domain \cite{bongard-etal-2022-legal}, task 5 of SemEval-2024 was proposed \cite{held-habernal-2024-legalreasoning}. The objective of this task is to discern the accurate responses to legal inquiries in U.S. Civil Procedure, as posited by the organizers. The questions and answers adhere to a Multiple-choice question-answering model, with accompanying explanations provided to facilitate comprehension of the legal concepts associated with each question. We have also released the code on GitHub \footnote{\url{https://github.com/haricharan189/Semeval_task5}.}

We delve into the foundational paradigms of machine learning, specifically focusing on Supervised and Unsupervised Learning, to introduce innovative approaches and present a comprehensive comparative analysis. The explanation part of our dataset undergoes a two-level segment-wise summarization generated by T5 \cite{48643}, which is consistently utilized throughout our investigation. Within the framework of the supervised setup, we leverage a multi-level CNN fusion approach \cite{8827512}, integrating LSTM and GRU architectures. This amalgamation facilitates the extraction of ensemble feature representations from questions, answers, and summaries. Additionally, a one-dimensional CNN model \cite{jacovi-etal-2018-understanding}, is trained. We employ a manual grid search technique to determine the optimal threshold that maximizes the macro F1 score, contributing to the refinement of our model.

In the unsupervised setup, we delve into the acquisition of diverse word representations such as word2vec and Glove. The assessment involves computing the similarity between question-answer pairs and answer-summary pairs, employing combinations like Glove-cosine, transformer embedding-cosine, transformer embedding-euclidean and word2vec-cosine. Notably, the best-performing supervised model achieved a macro F1 score of 66 \% on the development set and 49.6 \% on the test set. In contrast, the unsupervised approach yielded scores of 62 \% (development) and 52.3 \% (test). This outcome highlights a nuanced challenge related to generalization on the test set, prompting further exploration into the intricacies of model adaptability and robustness.

\section{Background}
The dataset provided by the organizers comprises three sets: Train Set, Dev Set, and Test Set, containing 666, 84, and 98 data points, respectively. Within the training and dev sets, each entry includes fields such as Question, Answer, Explanation, Label (with values of 0 or 1), Analysis, and Complete-Analysis providing a detailed examination. The test set, on the other hand, only consists of Question, Answer, and Explanation. The Label, when equal to 1, signifies a correct answer, while 0 denotes an incorrect one. The Explanation field provides context and background details for each question.

\begin{table}[!h]
\centering
\begin{tabular}{|l|p{4cm}|}
\hline
Field & Text \\
\hline
Explanation & The most basic point to understand about supplemental jurisdiction ........ on this basic purpose of Ârticle 1367(a). \\
Question & This and that. Garabedian, ........... are treated fairly. \\
Answer & has constitutional authority ............... under Ârticle 1367(a).  \\
Label & 0 \\
Analysis & Here, the Ârticle 1983 claim ............ Amendment claim.  \\
Complete analysis & This is pretty straightforward ............ D is the best choice here.  \\
\hline

\end{tabular}
\caption{Sample data-point from Train Set.}
\label{tab:accents}
\end{table}

\section{Related Works}
Legal texts pose a unique challenge for pre-trained transformers \cite{46201} due to the inclusion of specialized terminology not commonly used in everyday language . As a result, leveraging pre-trained models like BERT \cite{devlin-etal-2019-bert}, RoBERTa \cite{liu2019roberta}, and others becomes essential by training them on legal corpora to enhance their understanding of legal terminologies. Notable examples of transformers tailored for legal contexts include InLegalBERT \cite{10.1145/3594536.3595165}, Legal-RoBERTa \cite{geng2021legal}, and similar models.

Fine-tuning transformers, such as Legal-BERT \cite{chalkidis-etal-2020-legal}, on available legal data has been proposed as an effective strategy to improve performance on test sets, as suggested by Bongard et al. (2022) \cite{bongard-etal-2022-legal}. This approach capitalizes on domain-specific knowledge encoded during pre-training, enhancing the model's proficiency in handling legal language nuances.

In the domain of Legal Question Answering (LQA), recent works have extensively discussed significant advancements and challenges. The comprehensive review by \citeauthor{10.1016/j.cosrev.2023.100552} provides insights into the key works in LQA, outlining challenges and proposing future research directions. Louis et al. (2023) \cite{louis2023interpretable} shed light on the limitations of existing Large Language Models (LLMs) in Legal Question Answering, emphasizing the need for interpretability.

\section{System Overview}
 Transformers like T5, as demonstrated in the work of \cite{48643}, exhibit high efficiency in producing summaries for lengthy paragraphs. In this study, T5 was employed to generate segment-level summaries on explanation column using a two-step approach. The initial summary was created from the original text, with a segment length of 1000 tokens. These segment-wise summaries were then concatenated with spaces in between to form the first summary. Subsequently, the second summary was generated from the first summary, employing a segment length of 300 tokens, and similarly concatenated to provide a comprehensive summary of the input text. These summaries were used for further applications in place of explanation. Segment wise summary approach can be visualized as follows:
\begin{figure}[H]
      \centering
      \includegraphics[width=78mm]{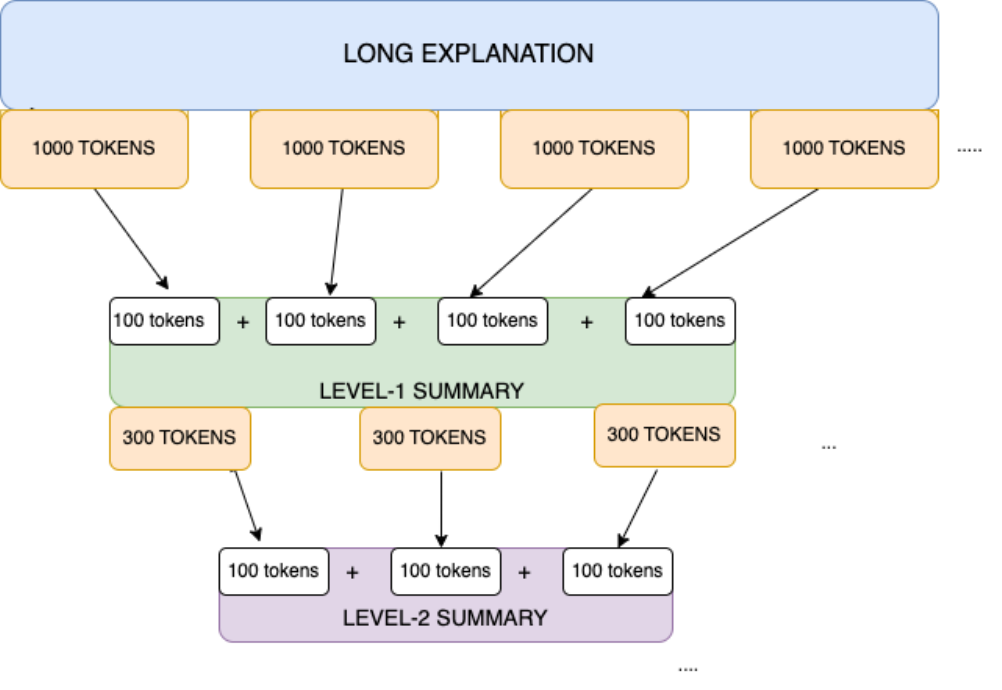}
      \caption{Segment wise summary}
      \label{fig:fig1}
      \end{figure}
      
\subsection{\large Supervised Models}

\subsubsection{Multi-Level Approach}
Following the generation of summaries, we employed the Legal-Bert transformer to extract embeddings from the question, answer, and summary columns. Each Legal-Bert output consists of a 768-dimensional vector, resulting in tensors of shape (number of data points, 768) for each dataset. Subsequently, we executed the following steps:

1.The tensor underwent a series of transformations through three consecutive 1-dimensional CNN layers, with ReLU activation functions \cite{10.5555/3104322.3104425}, and Adaptive max-pooling applied at each step. At each pooling layer, the output was reduced to 100 dimensions. The kernel size and padding were linearly increased, as depicted in the Figure 2.

2.The outputs from the first and second pooling layers were concatenated, yielding a first-level concatenated feature embedding of 200 dimensions.

3.This first-level output was then merged with the output from the third pooling layer to obtain a second-level concatenated embedding with 300 features.

4. Concurrently, the Legal-Bert embeddings were fed into Bi-GRU \cite{chung2014empirical} and Bi-LSTM \cite{10.1162/neco.1997.9.8.1735} models, resulting in 100 features from each. These features were concatenated.

5. The final multi-level feature representation was achieved by concatenating the second-level features with those from the GRU-LSTM models, resulting in a 500-dimensional vector. This process was applied to the question, answer, and summary, culminating in an exhaustive 1500-dimensional representation of the training data.

 \begin{figure}[H]
      \centering
      \includegraphics[width=80mm]{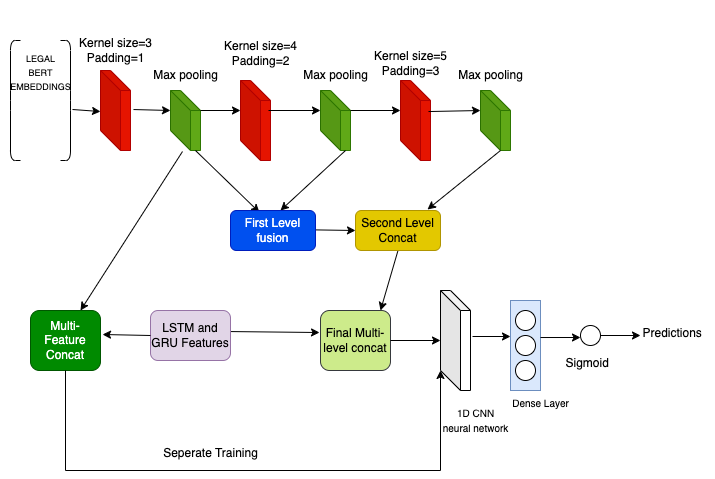}
      \caption{Multi Level fusion}
      \label{fig:fig2}
\end{figure}

\subsubsection{Multi-Feature Approach}

In this approach, the output of the first pooling layer was directly concatenated with the GRU-LSTM features, resulting in 300 features per entity, and hence, a 900-dimensional representation of the training data.

\textbf{Training and custom sigmoid layer:} To conduct a comparative analysis, we trained separate models using both multi-level and multi-feature representations. In each case, we employed a 1-dimensional CNN architecture implemented in TensorFlow, featuring a kernel size of 3 and 32 filters. Following max pooling, the resulting output was flattened and fed into a dense layer comprising 128 neurons. Finally, to enhance the variability of the probability distribution in the predictions, we introduced a custom Lambda layer. This layer subtracts the mean of the input tensor from each element and subsequently applies the sigmoid activation function.\\
\[
\text{{f}}(x) = y = \text{sigmoid}(x-\mu) \qquad (1)
\]
where \textmu \ is the mean of x \\ 
\textbf{Grid search and predictions:} Following the generation of probability vectors for the development set, we utilized manual grid search to determine the optimal threshold for classifying correct answers, aiming to maximize the macro-F1 score. Subsequently, the threshold associated with the highest F1 score on the development set was applied to make predictions on the test set

\subsection{\large Unsupervised Models}
\subsubsection{Word2Vec-Cosine system}
Word2Vec embeddings, as described in \cite{44876}, were extracted for the question, answer, and summary columns. A window size of 7 and a vector size of 5 were utilized for each word. Cosine similarities were computed between question-answer pairs and answer-summary pairs. The prediction was based on the mean of these similarities.

During evaluation, it was observed that in cases where the difference between the highest and second-highest similarity scores for a question was minimal, the answer with the second-highest similarity often turned out to be the correct answer. Consequently, a refinement was implemented: if the disparity between the highest and second-highest similarity scores was small, the answer with the second-highest similarity was labeled as 1, while the remaining answers were labeled as 0. This adjustment yielded improved results in such scenarios. A threshold of 0.0005 was used in this case after optimization on train and dev set.
\RestyleAlgo{ruled}
\begin{algorithm}
  \SetAlgoLined
  \caption{Word2Vec Similarity-based Labeling}\label{algorithm}
  
  \KwData{Word2Vec embeddings for question, answer, and summary columns}
  \KwResult{Labels for answers based on similarity scores}
  
  \For{each question}{
    max\_id= highest similarity score\;
    second\_max\_id = second-highest similarity score\;
    
    \If{$\lvert \text{similarity}[max\_id] - \text{similarity}[second\_max\_id] \rvert \leq 0.0005$}{
      Label[second\_max\_id] = 1\;
      Label the remaining answers as 0\;
    }\Else{
      Label[max\_id] = 1\;
      Label the remaining answers as 0\;
    }
  }
\end{algorithm}

\subsubsection{GloVE-Cosine system}
In contrast to the Word2Vec-Cosine approach, the methodology now incorporates GloVE embeddings as opposed to Word2Vec embeddings, leveraging the GloVE model proposed by Pennington et al. in 2014 \cite{pennington-etal-2014-glove}. Despite this shift, the overarching algorithm for label assignment remains unaltered, ensuring continuity and comparability with the Word2Vec-Cosine approach discussed in the preceding section.

\subsubsection{Transformer embeddings-Cosine system and Transformer embeddings-Euclidean system}
We utilized the Deberta model \cite{he2021deberta} trained on legal texts, specifically "LambdaX-AI/legal-deberta-v1," accessible on Hugging Face  \cite{wolf2020huggingfaces}. This model provided embeddings of questions, answers, and summaries, each represented by vectors of size 1536. We employed both cosine similarity and Euclidean distance metrics for label assignment.

For cosine similarity, the algorithm remained straightforward: answers with higher cosine similarity scores were assigned labels accordingly.

However, in the case of Euclidean distance, a slightly different approach was employed. The answer with the minimum distance was initially assigned a label of 1. Subsequently, if the difference between the minimum distance and the second minimum distance was less than a predefined threshold which is 0.8 in this case, the answer associated with the second minimum distance was labeled 1 instead, replacing the initial assignment.

\section{Experimental Setup}
We utilized Google Colab for training and testing our models, taking advantage of the T4 GPU provided by the platform.

\subsection{Supervised Models}

The Multi-feature concatenation method involved the integration of 900 features, while the Multi-level approach incorporated 1500 features. Both methodologies underwent training for 15 epochs with a batch size of 32. The optimization algorithm chosen was "Adam" \cite{kingma2017adam}, employing a learning rate set to 0.001.

\subsection{Unsupervised Models}

Word2Vec and GloVe embeddings were both generated with an embedding size of 5. However, there were differences in the window length used during training: for Word2Vec embeddings, a window length of 7 was utilized, while GloVe embeddings were trained with a window length of 10. In the case of GloVe, the training process spanned 30 epochs, employing a learning rate of 0.05 to optimize the model parameters.
These values of hyper-parameters were arrived after experimentation with several other values. 

\section{Results}

The performance metrics of our models on the test set and development set are presented in Table 2, where "Acc" represents accuracy and "F1" denotes the macro F1 score. Notably, our model demonstrated strong performance on the development set. However, it is worth mentioning that the performance on the test set was comparatively lower. It is important to highlight that our top-performing model utilizes an unsupervised approach leveraging Word2Vec embeddings and cosine similarity. Despite the varying performance, most of our models consistently outperformed the baseline.

\begin{table}[h]
\begin{tabular}{|p{2cm}|c|c|c|c|}
\hline
\multicolumn{5}{|c|}{Model Performance on Dev and Test set} \\
\hline
\multirow{2}{*}{Model} & \multicolumn{2}{c|}{Dev Set} & \multicolumn{2}{c|}{Test set} \\
\cline{2-5}
& Acc & F1 & Acc & F1 \\
\hline
Baseline & 0.798 & 0.444  &  \textbf{0.7449} &  0.4269 \\
\hline
Multi-level approach &  0.74 &  0.65 &  0.4898 &  0.4102 \\
\hline
Multi-Feature approach &  \textbf{0.81} &  \textbf{0.66} & 0.6224 &  0.4966 \\
\hline
Word2vec-cosine &  0.71 &  0.62 &  0.6429 & \textbf{0.5238}\\
\hline
\textit{Word2vec-cosine without replacement} &  0.62 &  0.56 & 0.6020 &  0.5072 \\
\hline
\textit{GloVE-cosine} &  0.64 &  0.56 & 0.6020 &  0.4694 \\
\hline
Transformer-cosine &  0.60 &  0.46 &  0.5612 &  0.4150 \\
\hline
\textit{Transformer-euclidean} &  0.60 &  0.46 &  0.5816 &  0.4421 \\
\hline
\textit{Transformer-manhattan} &  0.62 &  0.49 &  0.5612 &  0.4149 \\
\hline
\end{tabular}
\caption{Performance comparison of all our models}
\label{tab:mytable}
\end{table}

\begin{table}[h]
  \centering
    \begin{tabular}{|c|c|c|}
    \hline
    \multicolumn{3}{|c|}{Training Set Counts:} \\
    \hline
    Higher score & R/W & Count \\
    \hline
    Q     & R     & 143 \\
    Q     & W     & 81 \\
    S     & R     & 284 \\
    S     & W     & 158 \\
    \hline
    \multicolumn{3}{|c|}{Development Set Counts:} \\
    \hline
    Higher score & R/W & Count \\
    \hline
    Q     & R     & 11 \\
    Q     & W     & 14 \\
    S     & R     & 41 \\
    S     & W     & 18 \\
    \hline
    \end{tabular}%
    \caption{Distribution of right (R) and wrong (W) predictions before replacement}
  \label{tab:combined}%
\end{table}%

Analysis from Table 2 reveals a notable enhancement in model performance with the replacement of the second-best answer. The subsequent comparison, illustrated in Tables 3 and 4, highlights the impact of this replacement on the Wav2Vec-cosine model's results on both the training and development sets, considering the influence of two distinct similarity scores. Specifically, 'Q' signifies instances where the Question-Answer similarity surpasses the Summary-Answer similarity, while 'S' denotes the reverse scenario. The predictions of models in italics were submitted in Post-evaluation period. 

Observing Tables 3 and 4, it becomes evident that the number of accurate predictions substantially increases in the development set, relative to its total size. In the Codalab leader-board we ranked 16 out of 21 teams, and in the overall laederboard we ranked 15 out 21 teams.

\begin{table}[h]
  \centering
    \begin{tabular}{|c|c|c|}
    \hline
    \multicolumn{3}{|c|}{Training Set Counts:} \\
    \hline
    Higher score & R/W & Count \\
    \hline
    Q     & R     & 144 \\
    Q     & W     & 80 \\
    S     & R     & 286 \\
    S     & W     & 156 \\
    \hline
    \multicolumn{3}{|c|}{Development Set Counts:} \\
    \hline
    Higher score & R/W & Count \\
    \hline
    Q     & R     & 14 \\
    Q     & W     & 11 \\
    S     & R     & 46 \\
    S     & W     & 13 \\
    \hline
    \end{tabular}%
    \caption{Distribution of right (R) and wrong (W) predictions after replacement}
  \label{tab:combined}%
\end{table}%

\section{Conclusion and Future scope}
The dataset presents challenges for models to grasp the intricate legal context, resulting in subpar performance of regular supervised models. Unsupervised models heavily rely on embeddings, but available transformers inadequately capture the dataset's nuances. These models operate under the assumption of at least one correct answer per question; however, instances where all answers were labeled as incorrect hindered unsupervised model performance.

Future endeavors entail amalgamating these models into a unified super model. This super model would aggregate predictions from various models to yield a singular final prediction, enhancing overall performance and addressing the limitations of individual approaches. An alternative strategy involves leveraging Siamese networks to learn similarity, addressing challenges encountered by unsupervised models when all answers for a particular question are labeled as incorrect (0). By employing Siamese networks, we believe that the model can effectively capture nuanced similarities between question-answer pairs, and provide better predictions. Exploring other kind of summarizers and using other transformers for summarization such BART \cite{lewis-etal-2020-bart} may also increase the overall performance of all the systems used in this paper. Data augmentation \cite{feng-etal-2021-survey} can also be implemented to get better Word2Vec and GloVE embeddings. 

\section*{Acknowledgements}

We would like to thank the organizers, reviewers and SemEval - 2024 Chairs for their valuable insights and helpful suggestions.

% Bibliography entries for the entire Anthology, followed by custom entries
%\bibliography{anthology,custom}
% Custom bibliography entries only
\bibliography{custom}

\end{document}